\documentclass{article}

 \usepackage{comment}
 \usepackage{caption} 

 \usepackage[dblblindworkshop, final]{neurips_2025}
\workshoptitle{Efficient Reasoning}

\usepackage[utf8]{inputenc} 
\usepackage[T1]{fontenc}    
\usepackage{hyperref}       
\usepackage{url}            
\usepackage{booktabs}       
\usepackage{amsfonts}       
\usepackage{nicefrac}       
\usepackage{microtype}      
\usepackage{xcolor}         

\usepackage{tikz}
\usetikzlibrary{shapes.geometric, arrows.meta, positioning, calc, fit}

\usepackage{algorithm}
\usepackage{algpseudocode}
\usepackage{float}       
\usepackage{placeins}    
\usepackage{caption}
\usepackage{tikz}
\usetikzlibrary{positioning,calc,arrows.meta}
\usepackage{float}

\usepackage{lineno}

\usepackage{booktabs}   
\usepackage{multirow}   
\usepackage{caption}    
\usepackage{siunitx}    
\usepackage{float}      

\usepackage{array}      
\usepackage{booktabs}   
\usepackage{amsmath}    
\usepackage{subcaption} 
\title{Chopping Trees: Semantic Similarity Based Dynamic Pruning for Tree-of-Thought Reasoning}

\author{
\textbf{Joongho Kim}$^{1,*,\ddagger}$ \quad
\textbf{Xirui Huang}$^{1}$ \quad
\textbf{Zarreen Reza}$^{1, \dagger}$ \quad
\textbf{Gabriel Grand}$^{2, \dagger}$ \\ 
$^{1}$Independent Researcher \quad $^{2}$Massachusetts Institute of Technology (MIT) \\
\texttt{joonghokim@google.com}
}

\makeatletter
\let\@oldmaketitle\@maketitle
\def\@maketitle{%
  \@oldmaketitle%
  \begingroup
  \renewcommand{\thefootnote}{\fnsymbol{footnote}}%
  \footnotetext[1]{Lead Author.}%
  \footnotetext[2]{Senior Authors.}%
  \footnotetext[3]{Now at Google.}%
  \endgroup
}
\makeatother

\begin{document}

\maketitle
\begin{abstract}

Tree-of-Thought (ToT) reasoning boosts the problem-solving abilities of Large Language Models (LLMs) but is computationally expensive due to \textbf{semantic redundancy}, where distinct branches explore equivalent reasoning paths. We introduce \textbf{Semantic Similarity-Based Dynamic Pruning (SSDP)}, a lightweight method that, to the best of our knowledge, is the first framework to integrate online semantic merging into parallelized tree search, enabling the clustering and pruning of redundant steps in real time. Across reasoning benchmarks, including GSM8K and MATH500, SSDP achieves up to a \textbf{2.3x speedup} over state-of-the-art tree-search baselines while maintaining competitive accuracy (typically within 5\% of the strongest baseline) and reducing the number of explored nodes by \textbf{85–90\%}, demonstrating a practical approach to efficient, scalable LLM reasoning. \noindent{The implementation of SSDP is publicly available at} \url{https://github.com/kimjoonghokim/SSDP}.

\end{abstract}

\section{Introduction}
\label{intro}
Tree-of-Thought (ToT) search strategies \citep{yao2023tree} enable Large Language Models to explore multiple reasoning paths beyond linear Chain-of-Thought prompting \citep{wei2022chain}, improving performance on complex problems requiring backtracking or strategic lookahead. However, exploring vast reasoning trees remains computationally expensive, with many branches redundantly exploring semantically equivalent reasoning paths.

We introduce \textbf{Semantic Similarity-Based Dynamic Pruning (SSDP)}, a lightweight method that identifies and merges semantically similar reasoning branches into a single representative "hypernode" during search. SSDP requires only a reward model and its core semantic merging logic to dynamically prune the search space, without dataset-specific fine-tuning. To the best of our knowledge, SSDP is the first framework that integrates online semantic merging, that is, merging semantically similar reasoning paths dynamically during inference rather than after completion, into parallelized tree search to cluster and prune redundant steps in real time.

Across challenging benchmarks like GSM8K and MATH500, SSDP achieves up to a \textbf{2.3x speedup} over strong tree-search baselines while maintaining comparable accuracy. It also dramatically reduces the search space, exploring on average \textbf{85-90\% fewer nodes}, demonstrating that targeting semantic redundancy is an effective and practical strategy for scalable LLM reasoning.

Our contributions are threefold:
\begin{enumerate}
    \item We propose SSDP, a novel and lightweight pruning method that leverages semantic similarity to make tree-based search more efficient and, to the best of our knowledge, is the first framework of its kind used for inference time scaling.
    \item We provide extensive empirical evidence showing that SSDP substantially reduces inference time and computational load across multiple models and benchmarks, without a significant trade-off in accuracy.
    \item We demonstrate that explicitly targeting semantic redundancy is a highly effective strategy for optimizing LLM reasoning at inference time, paving the way for more scalable and practical applications of advanced reasoning frameworks.
\end{enumerate}

\section{Related Work}
\label{rel_work}
Our work lies at the intersection of tree search for LLM reasoning, efficiency-focused pruning and merging, and semantic similarity for reasoning.

\paragraph{Tree Search for LLM Reasoning.}
Linear strategies such as Chain-of-Thought (CoT) \citep{wei2022chain} and Self-Consistent CoT \citep{wang2022} improved multi-step reasoning but explored limited paths. Tree-of-Thoughts (ToT) \citep{yao2023tree} expanded exploration using sampling, beam search, and MCTS \citep{cobbe2021, hao2023, wan2024b}, aligning with the "inference scaling law" \citep{MLSYS2024_5321b1da, lin2024awqactivationawareweightquantization, fu2024breaksequentialdependencyllm, cai2024medusasimplellminference}. Dynamic Parallel Tree Search (DPTS) \citep{ding2025dynamicparalleltreesearch} improved parallel efficiency but still explored many redundant states. \textbf{SSDP addresses this by pruning redundant paths before they are generated.} Recent work has acknowledged that tree-based methods continue to suffer from computational overheads \citep{rufail2025clearcontrastingtextualfeedback}, reinforcing the need for efficiency-focused optimizations like SSDP.

\paragraph{Pruning and Merging for Efficiency.}
Pruning improves reasoning even in linear CoT \citep{zhao2025pruningimprovereasoningrevisiting}. Methods like FETCH \citep{wang2025dontlosttreesstreamlining} merge semantically similar states but require fine-tuned models. \textbf{SSDP provides a lightweight, general alternative}, using only a small reward model and semantic merging to achieve large efficiency gains without task-specific tuning.

\paragraph{Semantic Similarity in Verification.}
Semantic Self-Consistency \citep{knappe2025semanticselfconsistencyenhancinglanguage} clusters completed reasoning chains post-hoc. In contrast, \textbf{SSDP integrates semantic merging during tree growth}, shaping the search process itself to make reasoning faster and more efficient.

\section{Methodology}
\label{method}

\begin{figure}[H]
\centering

\begin{subfigure}[t]{0.42\textwidth}
  \centering
  \resizebox{0.95\linewidth}{!}{
    \begin{tikzpicture}[
      node/.style = {rectangle, draw=black!70, rounded corners, font=\footnotesize, minimum width=28mm, align=center},
      arrow/.style = {->, >=Stealth, semithick}
    ]
    \node[node] (start) {Init: root $r$, \\ $Q \leftarrow [r]$};
    \node[node, below=8mm of start] (pop) {Pop $n$ from $Q$};
    \node[node, below=8mm of pop] (expand) {Expand($n,k$)\\generate $\mathcal{C}$};
    \node[node, below=8mm of expand] (merge) {Score \& prune\\MergeSimilar($\mathcal{R},\sigma$) $\rightarrow \mathcal{M}$};
    \node[node, below=8mm of merge] (insert) {Insert $\mathcal{M}$ into $Q$};
    \node[node, below=8mm of insert] (done) {Return best candidate};

    \draw[arrow] (start) -- (pop);
    \draw[arrow] (pop) -- (expand);
    \draw[arrow] (expand) -- (merge);
    \draw[arrow] (merge) -- (insert);
    \draw[arrow] (insert) -- (done);

    \draw[arrow] (insert.west) .. controls +(-20mm,0) and +(-20mm,0) .. (pop.west);

    \node[font=\scriptsize, align=left, right=4mm of merge] {
      \textbf{Notes:}\\
      \scriptsize Prune if $S(c)<\tau$\\
      Representative = best-score / centroid
    };
    \end{tikzpicture}
  }
  \caption{\textbf{Figure 1a:} SSDP flowchart.}
  \label{fig:ssdp-vertical-side}
\end{subfigure}%
\hfill
\begin{subfigure}[t]{0.58\textwidth}
  \centering
  \resizebox{0.95\linewidth}{!}{%
    \begin{tikzpicture}[
      scale=1,
      every node/.style={font=\small},
      treenode/.style={circle, draw=black!80, minimum size=11mm, inner sep=0.8pt, align=center, fill=white},
      leaf/.style={circle, draw=black!80, minimum size=9.5mm, inner sep=0.6pt, align=center},
      repnode/.style={circle, draw=black!90, thick, fill=yellow!20, minimum size=12mm, inner sep=0.6pt, align=center},
      simline/.style={dashed, gray!60, semithick, -{Stealth[length=2.2mm,width=1.2mm]}},
      mergearrow/.style={very thick, black!80, -{Stealth[length=2.6mm,width=1.7mm]}},
      arr/.style={->, >=Stealth, semithick}
    ]

    \node[treenode] (r) at (0,1.8) {$r$};
    \node[treenode] (A) at (-5.2,0.2) {A\\s=0.82};
    \node[treenode] (B) at (0,0.2)    {B\\s=0.74};
    \node[treenode] (C) at (5.2,0.2)  {C\\s=0.80};

    \node[leaf] (A1) at (-6.4,-1.4) {A1\\s=0.77};
    \node[leaf] (A2) at (-4.0,-1.4) {A2\\s=0.79};
    \node[leaf] (B1) at (-1.6,-2.8) {B1\\s=0.65};
    \node[leaf] (B2) at ( 1.6,-2.8) {B2\\s=0.63};
    \node[leaf] (C1) at ( 4.0,-1.4) {C1\\s=0.75};
    \node[leaf] (C2) at ( 6.4,-1.4) {C2\\s=0.78};

    \node[repnode] (M1) at (-5.2,-2.8) {M$_1$\\s=0.74};
    \node[repnode] (M2) at (5.2,-2.8) {M$_2$\\s=0.69};

    \draw[arr] (r) -- (A);
    \draw[arr] (r) -- (B);
    \draw[arr] (r) -- (C);

    \draw[arr] (A) -- (A1);
    \draw[arr] (A) -- (A2);
    \draw[arr] (B) -- (B1);
    \draw[arr] (B) -- (B2);
    \draw[arr] (C) -- (C1);
    \draw[arr] (C) -- (C2);

    \draw[simline] (A1) to[out=0,in=180] node[midway, above, yshift=1mm, font=\scriptsize] {sim=0.92} (A2);
    \draw[simline] (C1) to[out=0,in=180] node[midway, above, yshift=1mm, font=\scriptsize] {sim=0.89} (C2);

    \draw[mergearrow] (A1) -- (M1);
    \draw[mergearrow] (A2) -- (M1);
    \draw[mergearrow] (C2) -- (M2);
    \draw[mergearrow] (C1) -- (M2);

    \node[anchor=north] (legend) at (0,-3.2) {
      \begin{tabular}{@{}c@{}}
        \textbf{Legend:} \\[3pt]
        \tikz{\node[treenode,minimum size=5mm, inner sep=0pt]{};} \ Node (score) \quad
        \tikz{\node[repnode,minimum size=5mm, inner sep=0pt]{};} \ Merged representative \\[3pt]
        \tikz{\draw[simline] (0,0) -- (7mm,0);} \ Semantic similarity \quad
        \tikz{\draw[mergearrow] (0,0) -- (7mm,0);} \ Merge $\rightarrow$ representative
      \end{tabular}
    };

    \end{tikzpicture}
  }
  \caption{\textbf{Figure 1b:} SSDP tree.}
  \label{fig:ssdp-horizontal-side}
\end{subfigure}
\label{fig:ssdp-two-panels}

\vspace{1mm}
\caption{
\textbf{Overview of SSDP.} 
(\textbf{a}) Flowchart of the SSDP algorithm showing its iterative expand–score–prune–insert loop. 
(\textbf{b}) Example search tree where semantically similar nodes are merged into representative clusters, 
reducing redundant exploration.}
\label{fig:ssdp-two-panels}
\end{figure}
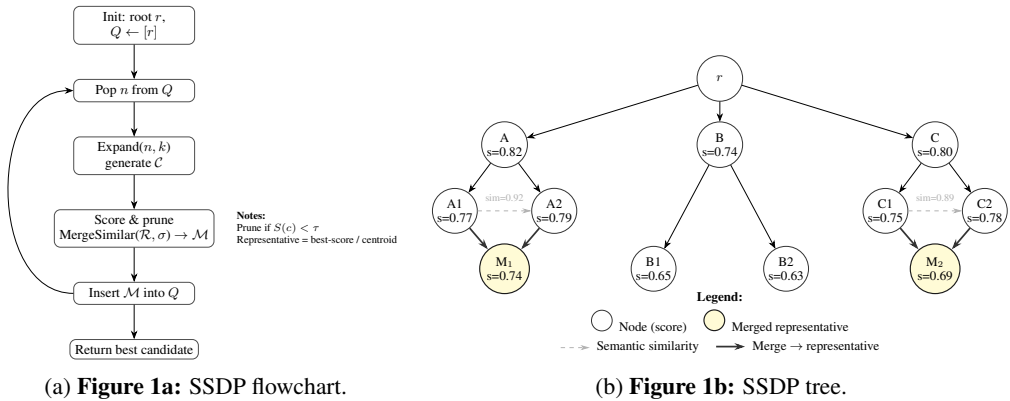

In this section, we present the design of \textbf{Semantic Similarity--based Dynamic Pruning (SSDP)}. 
We first describe the pruning-augmented tree search framework, then introduce the semantic similarity 
merging mechanism, detail the node-level scoring and pruning strategy, and finally present our search 
and stopping criteria. 

An overview of the SSDP process is shown in Figure~\ref{fig:ssdp-vertical-side} (flowchart) and 
Figure~\ref{fig:ssdp-horizontal-side} (tree example). Figure~\ref{fig:ssdp-vertical-side} illustrates the iterative tree 
expansion and pruning loop, while Figure~\ref{fig:ssdp-horizontal-side} visualizes how semantically similar nodes are merged 
into representative clusters during search.

\captionsetup[subfigure]{labelformat=empty} 

\subsection{Framework Overview}

SSDP is designed to improve tree-based LLM reasoning efficiency by pruning semantically redundant nodes. It builds upon the parallel tree search framework of DPTS \citep{ding2025dynamicparalleltreesearch}, inheriting dynamic expansion control. 

The search loop follows standard MCTS-style stages: selection of promising nodes, expansion to generate children, evaluation using a reward model, semantic merging and pruning (our main contribution), and backpropagation of node scores. This process reduces redundant reasoning paths and leverages parallelism for efficient computation.

The search loop proceeds as follows:

\begin{enumerate}
    \item \textbf{Selection:} The framework first selects up to $k$ promising leaf nodes from the search frontier to expand in parallel. This selection is guided by an Upper Confidence Bound (UCB) policy, defined by \citep{ding2025dynamicparalleltreesearch} as:
 {\footnotesize
\[
\mathrm{UCB}(n) = \frac{Q_n}{N_n} + w \, \sqrt{1+N_{\mathrm{parent}(n)}} \cdot \frac{\phi(n)}{1+N_n}
\]
}
    where $Q_n$ is the cumulative reward, $N_n$ is the visit count, $\phi(n)$ is a node's intrinsic score from a reward model, and $w$ is an exploration weight.

    \item \textbf{Expansion and Evaluation:} For each selected node, the base LLM generates $b$ candidate children via temperature sampling. Each new child node is then evaluated by a reward model to obtain its score.

    \item \textbf{Semantic Merging:} Before the newly generated children are added to the search frontier, they undergo our semantic merging process. As detailed in Section~\ref{sec:ssdp-merge}, this step identifies and consolidates semantically equivalent nodes into a single representative. This fundamentally reduces the effective branching factor of the search tree, preventing the system from wasting resources on redundant reasoning paths.

    \item \textbf{Backpropagation:} Finally, the reward from the newly evaluated nodes is propagated up the tree to update the statistics ($Q_u, N_u$) of their ancestors, following the standard MCTS procedure.
\end{enumerate}

This process is further accelerated by parallel execution and generic efficiency mechanisms, such as early stopping for unpromising rollouts.

\subsection{Semantic Merging Module}\label{sec:ssdp-merge}
The core innovation of SSDP is its online semantic merging module, which prunes the search tree at each expansion step. This process consists of three stages: embedding, clustering, and representative selection.

1. \textbf{Embedding Generation:} Each child node $c$ is encoded into a dense vector $s(c) = \mathcal{E}(\text{text}(c))$ using a frozen sentence-transformer. 
The resulting embeddings are $\ell_2$-normalized to facilitate efficient cosine similarity calculations.

2. \textbf{Similarity-Based Clustering:} Sibling nodes with cosine similarity above threshold $\tau$ are grouped into clusters of semantically equivalent nodes.

3. \textbf{Representative Selection:} From each cluster $C$, only the node with highest reward score $\phi(c)$ is retained; all others are pruned to free memory. The unique representatives $\{c^*_1, c^*_2, \dots\}$ are added to the frontier. 
\[
c^* = \arg\max_{c \in C} \phi(c)
\]
All other nodes within the cluster are discarded, and their associated KV caches are immediately freed to reduce memory overhead. Only the set of unique representatives $\{c^*_1, c^*_2, \ldots\}$ is added to the search frontier. This aggressive, online pruning strategy is the primary mechanism through which SSDP reduces the number of nodes explored and accelerates the time to solution.

\section{Experimental Setup}
\label{sec:exp_setup}

Our SSDP implementation builds on the official Dynamic Parallel Tree Search (DPTS) codebase \citep{ding2025dynamicparalleltreesearch}, adding semantic merging during node expansion while retaining other architectural components for fair comparison.

\subsection{Models, Datasets, and Baselines}

\textbf{Models.} We evaluate four open-source, instruction-tuned models from the Llama and Qwen families: \texttt{Llama-3.1-8B-Instruct} \citep{meta2024llama31}, \texttt{Llama-3.2-3B-Instruct}, \texttt{Qwen2.5-1.5B-Instruct}, and \texttt{Qwen2.5-7B-Instruct} \citep{qwen2024qwen2}. These were chosen to ensure our findings are robust across different model architectures and scales.

\textbf{Datasets.} Experiments were conducted on two standard multi-step reasoning benchmarks: \textbf{GSM8K} \citep{cobbe2021} and \textbf{MATH500} \citep{hendrycks2021math}. These datasets have canonical scoring procedures, enabling direct comparison with prior work.

\textbf{Baselines.} We compare SSDP against four widely-used search methods to quantify its impact: (1)~\textbf{DPTS}~\citep{ding2025dynamicparalleltreesearch}, the state-of-the-art parallel search framework; (2)~\textbf{MCTS}, a standard Monte Carlo Tree Search implementation for LLM reasoning \citep{yao2023tree}; (3)~\textbf{Best-of-N}, which samples independent CoT traces and selects the best result \citep{cobbe2021}; and (4)~\textbf{Beam Search}, which maintains a fixed-size beam of the most promising reasoning paths \citep{yao2023tree}.

\subsection{Evaluation Protocol}

For a fair comparison, all methods were run with identical prompts, reward models, and matched computational budgets (e.g., maximum expansions). Our primary evaluation metrics are:
\begin{itemize}
    \item \textbf{Accuracy:} Canonical exact-match scoring for each dataset.
    \item \textbf{Time (s):} Average inference time required to process a sample in the dataset.
\end{itemize}

As a secondary analysis, we also measured the average number of nodes generated and explored per sample for DPTS and SSDP using Qwen-2.5 1.5B. This secondary experiment on Qwen-2.5 1.5B examines node exploration efficiency to help explain observed performance differences (Table ~\ref{tab:nodes_combined}).



\section{Results}
\label{results}


\begin{table}[ht]
\centering
\captionsetup{font=footnotesize} 
\footnotesize 
\caption{Comparisons across search algorithms on LLM reasoning tasks. (Higher score is better).}
\label{tab:results}
\begin{tabular}{llcccccc}
\toprule
Model & Method & \multicolumn{3}{c}{MATH500} & \multicolumn{3}{c}{GSM8K} \\
 & & Acc. (\%) & Time (s) & Acc. (\%) & Time (s)\\
\midrule
\multirow{5}{*}{Qwen-2.5 1.5B} & MCTS\textsuperscript{\dag}     & 56.6 & 117.37 & 75.1 & 73.28 \\
                               & Best-of-N\textsuperscript{\dag} & 52.6 & 89.87 & 70.1 & 33.37 \\
                               & Beam\textsuperscript{\dag}      & 52.4 & 104.58 & 71.5 & 41.27 \\
                               & DPTS      & 58.6 & 37.4 & 80.9 & 14.7 \\
                               & SSDP      & 58.24 & \textbf{19.01} & 75.54 & \textbf{6.43} \\
\midrule
\multirow{5}{*}{Qwen-2.5 7B}   & MCTS\textsuperscript{\dag}      & 75.2 & 121.46 & 89.6 & 79.68 \\
                               & Best-of-N\textsuperscript{\dag} & 71.6 & 91.29 & 88.2 & 34.89 \\
                               & Beam\textsuperscript{\dag}      & 72.4 & 106.89 & 86.7 & 36.49 \\
                               & DPTS      & 76.3 & 44.5 & 89.5 & 19.9 \\
                               & SSDP      & 75.56 & \textbf{22.35} & 87.73 & \textbf{7.08} \\
\midrule
\multirow{5}{*}{Llama-3 3B}    & MCTS\textsuperscript{\dag}      & 48.6 & 111.80 & 64.0 & 57.19 \\
                               & Best-of-N\textsuperscript{\dag} & 46.4 & 91.34 & 57.1 & 27.27 \\
                               & Beam\textsuperscript{\dag}      & 45.2 & 104.36 & 58.4 & 28.27 \\
                               & DPTS      & 56.8 & 39.1 & 57.8 & 9.3 \\
                               & SSDP      & 52.47 & \textbf{15.54} & 56.52 & \textbf{4.3} \\
\midrule
\multirow{5}{*}{Llama-3 8B}    & MCTS\textsuperscript{\dag}      & 54.2 & 143.36 & 69.5 & 69.74 \\
                               & Best-of-N\textsuperscript{\dag} & 49.8 & 122.63 & 67.6 & 33.48 \\
                               & Beam\textsuperscript{\dag}      & 49.6 & 142.21 & 68.3 & 34.51 \\
                               & DPTS      & 61.9 & 49.8 & 62.4 & 12.8 \\
                               & SSDP      & 60.19 & \textbf{20.37} & 61.83 & \textbf{5.67} \\
\bottomrule
\end{tabular}

\vspace{1mm}
{\footnotesize
\dag\ Results were obtained from the DPTS paper \citep{ding2025dynamicparalleltreesearch}.}
\end{table}

Across the evaluated methods and benchmarks, SSDP consistently reduced inference latency while preserving the quality of the final answer. Table~\ref{tab:results} reports the accuracy and inference times of the models for each method. In terms of correctness, SSDP attains parity with the strongest baselines: across GSM8K the SSDP accuracies closely track DPTS, Beam and Best-of-N, and on MATH500 SSDP matches or outperforms several non-DPTS baselines in multiple model settings. In summary, we do not observe systematic degradation in the accuracy of the final answer attributed to semantic merging.


The runtime improvements from SSDP are substantial and consistent. Aggregating across the four methods and two data sets (Table~\ref{tab:ssdp_speedups}), SSDP reduces the total inference time by approximately \(2.3\times\), \(5.2\times\) , \(5.9\times\), and \(7.7\times\), when compared to DPTS, Best-of-N, Beam and MCTS respectively. These ratios indicate that SSDP eliminates a large fraction of redundant model computation, yielding a significantly faster completion for every search strategy tested.




\begingroup
\setlength{\textfloatsep}{8pt} 
\setlength{\intextsep}{6pt}    
\setlength{\abovecaptionskip}{3pt} 
\setlength{\belowcaptionskip}{3pt} 

\begin{table}[htbp]
\centering
\captionsetup{font=footnotesize}
\caption{Aggregate SSDP speedups}
\label{tab:ssdp_speedups}
\setlength{\tabcolsep}{6pt}
\begin{tabular}{lc}
\toprule
\textbf{Baseline} & \textbf{Implied speedup (×)} \\
\midrule
DPTS         & 2.26 \\
Best-of-N    & 5.20 \\
Beam         & 5.94 \\
MCTS         & 7.68 \\
\midrule
\multicolumn{2}{l}{\textit{Model-family averages}} \\
Llama family & 5.87 \\
Qwen family  & 4.77 \\
\bottomrule
\end{tabular}
\end{table}
\endgroup

We also observe modest model-family differences: the proportional speedups are slightly larger for the Llama family than for Qwen in our runs 
(\(T_{\mathrm{baseline}} / T_{\mathrm{SSDP}} \approx 5.87\) for Llama vs.\ \(\approx 4.77\) for Qwen; Table~\ref{tab:ssdp_speedups}). The values are calculated by taking the mean SSDP speedup ratio against all four search methods and across both datasets for models belonging to the same family. This pattern is consistent with the intuition that models with higher per-token or per-step generation cost benefit more from avoiding redundant rollouts, although the magnitude depends on decoding and batching settings.





\begingroup
\setlength{\textfloatsep}{6pt} 
\setlength{\intextsep}{6pt}    
\setlength{\abovecaptionskip}{3pt} 
\setlength{\belowcaptionskip}{3pt} 

\begin{table}[htbp]
\centering
\captionsetup{font=footnotesize}
\scriptsize
\setlength{\tabcolsep}{6pt}

\caption{Average nodes generated and explored per sample (Qwen2.5-1.5B)}
\label{tab:nodes_combined}
\resizebox{0.7\linewidth}{!}{%
  \begin{tabular}{lcccc}
    \toprule
    Method
    & \multicolumn{2}{c}{\textbf{MATH500}}
    & \multicolumn{2}{c}{\textbf{GSM8K\textsuperscript{\dag}}} \\
    \cmidrule(lr){2-3} \cmidrule(lr){4-5}
    & Nodes gen. & Nodes expl. & Nodes gen. & Nodes expl. \\
    \midrule
    DPTS & 214.4 & 53.3 & 79.1 & 19.5 \\
    SSDP & \textbf{31.1} & \textbf{11.5} & \textbf{9.4} & \textbf{4.8} \\
    \bottomrule
  \end{tabular}%
}

\vspace{1mm}
{\footnotesize
\textsuperscript{\dag}\,Experiment was performed on a 500 sample subset of GSM8K.
}

\setlength{\tabcolsep}{6pt}
\end{table}
\endgroup

Diagnostic counts corroborate the mechanistic source of these gains. Table~\ref{tab:nodes_combined} shows that, for Qwen-2.5-1.5B, SSDP generates far fewer candidates on average (MATH500: 31.1 vs.\ 214.4; GSM8K: 9.4 vs.\ 79.1) and explores substantially fewer nodes per sample (MATH500: 11.5 vs.\ 53.3; GSM8K: 4.8 vs.\ 19.5). The reduction in generated and explored nodes aligns with the observed inference time savings and supports the interpretation that sibling-level semantic merging prevents repeated model invocations on near-duplicate continuations.

Taken together, these results demonstrate that SSDP is an effective, low-cost augmentation to Tree-of-Thought reasoning processes: it achieves large, robust latency reductions by pruning semantic redundancy while maintaining final-answer accuracy. 

Based on this analysis, we selected $\tau = 0.75$ as the default threshold for our experiments, as it provides a strong balance between efficiency and accuracy. However, the threshold can be adjusted depending on whether the user prioritizes speed (lower $\tau$) or accuracy (higher $\tau$).

\subsection{Ablation Study}
\begin{figure}[H]
    \centering
    \includegraphics[width=0.8\linewidth]{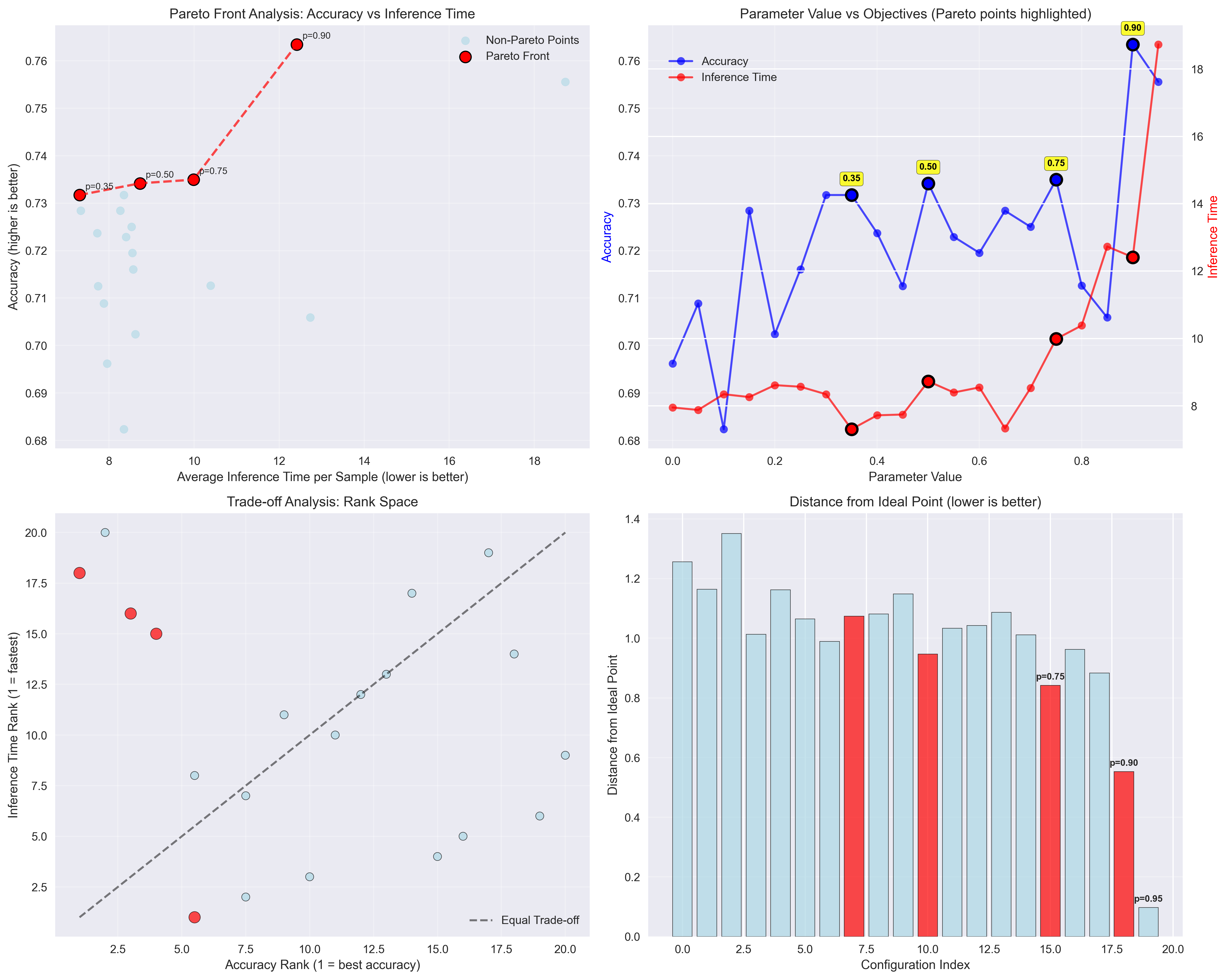}
    \caption{Pareto Front Analysis on Similarity Threshold $\tau$ }
    \label{fig:placeholder}
\end{figure}
\FloatBarrier
We conducted an ablation study to analyze the impact of the similarity threshold $\tau$ on the performance of SSDP. The experiments were performed on the GSM8K dataset using Qwen-2.5 1.5B, varying $\tau$ from 0 to 1 in steps of 0.05. For each threshold value, we measured both accuracy and inference time to understand the trade-off between speed and performance.

The results of this analysis reveal four threshold values that lie on the Pareto front, representing optimal trade-offs between speed and accuracy: 0.35, 0.5, 0.75, and 0.9. Specifically:
\begin{itemize}
    \item $\tau = 0.35$: fastest configuration while still achieving some accuracy gains.
    \item $\tau = 0.5$ and $0.75$: balanced configurations offering good trade-offs between speed and accuracy.
    \item $\tau = 0.9$: highest accuracy but slower inference.
\end{itemize}

\section{Limitations}
\label{sec:limit}

\paragraph{Semantic similarity does not imply equivalence (and vice versa).}
SSDP assumes that high embedding similarity between intermediate steps indicates functional redundancy, yet semantically similar text can diverge logically, and conversely, two phrasing-distant steps can be logically equivalent. This mismatch risks (i) pruning a step that later enables a distinct, correct derivation, or (ii) retaining a near-duplicate that appears dissimilar in embedding space. Practical mitigations include conservative thresholds, verifier-aware gating, and deferred merging for uncertain clusters.

\paragraph{Evidence concentrated on math; broader validation needed.}
Empirical results are centered on math-style reasoning (e.g., GSM8K, MATH500). While these benchmarks stress multi-step deduction, they may not capture the linguistic variability, world knowledge, or planning demands of other domains (code generation, multi-hop QA, commonsense, tool use). Generalization to such settings and across model sizes/families remains an open question requiring domain-specific studies and possibly task-adapted similarity encoders.

\paragraph{Similarity threshold (\texorpdfstring{$\tau$}{tau}) is hand-tuned.}
The current system relies on a single, manually selected \( \tau \) to decide when steps are “similar enough” to merge. This choice is dataset- and model-dependent, and small changes can shift pruning behavior noticeably. As a result, reported results may partially reflect the chosen threshold rather than an intrinsic property of the method; understanding performance across a range of \( \tau \) remains important.

\paragraph{Potential over-pruning of distinct reasoning.}
When two steps read alike, the procedure may collapse them even if they pursue materially different ideas. Such removals can narrow exploration and hide alternative derivations that would have succeeded later in the search. The extent of this effect is not fully quantified here, so accuracy losses attributable to over-pruning may be underreported.

\paragraph{Embedding backbone cost and sensitivity.}
The approach inherits the computational footprint and biases of the embedding model used to judge similarity. Encoder latency and memory add overhead to search time, and domain mismatches can distort which steps appear “close.” Performance and conclusions may therefore vary with the encoder choice, dimensionality, and preprocessing, which are not exhaustively analyzed in this work.

\section{Conclusion}
\label{conclusion}

We introduced Semantic Similarity Based Dynamic Pruning (SSDP), a lightweight method to address the critical inefficiency in Tree-of-Thought (ToT) reasoning caused by semantic redundancy. By integrating an online semantic merging module into a parallel search framework, SSDP identifies and prunes redundant reasoning paths in real-time without requiring complex fine-tuning. Our experiments across multiple models and benchmarks demonstrate that this approach yields substantial gains: SSDP achieves up to a \textbf{2.3x speedup} over the parallel-search DPTS baseline and even greater gains (5-8x) over other standard search methods while maintaining comparable accuracy. This efficiency is a direct result of reducing the search space by an average of \textbf{85-90\%}. By showing that a simple, general-purpose pruning mechanism can yield substantial performance gains, SSDP offers a practical path toward making inference-time scaling more efficient. It paves the way for more complex and deliberative AI reasoning to become not only more powerful but also scalable and practical for real-world applications.

\medskip





\newpage
\bibliographystyle{plain}
\bibliography{references}

\newpage
\appendix

\section{Technical Appendices and Supplementary Material}
\label{appexdix:technical}

\subsection{Reproducibility Statement}
\begin{itemize}
    \item \textbf{Code Availability:} All code used for inference and evaluation will be available at \url{https://github.com/kimjoonghokim/SSDP}.
    \item \textbf{Models:} The Llama 3 family of models has restricted access, but is made available by Meta on request. You can access them by requesting permission through the provided \href{https://huggingface.co/meta-llama}{Meta license}. The \href{https://huggingface.co/Qwen}{Qwen family} of models are publicly accessible and can be found on HuggingFace
    \item \textbf{Datasets:} Both MATH500 and GSM8K datasets used are publicly available through HuggingFace and are also accessible through our codebase.
   \item \textbf{Prompts:} All prompts were taken from the DPTS paper \citep{ding2025dynamicparalleltreesearch}.
\end{itemize}

\subsection{GPU Usage}
\label{appendix:gpu}
\begin{table}[h]
\centering
\setlength{\tabcolsep}{6pt}
\caption{GPUs used for code testing and running experiments.}
\renewcommand{\arraystretch}{1.3}

\begin{tabular}{@{} l c p{5.0cm} c @{}}
\toprule
\textbf{GPU Model} & \textbf{Memory (GB)} & \textbf{Usage Purpose} & \textbf{Approx. Hours Used} \\
\midrule
NVIDIA H200 SXM & 141 & Running experiments & 40 \\
NVIDIA A40      & 48  & Code testing and development & 120 \\
\bottomrule
\end{tabular}

\label{tab:gpus}
\end{table}

\subsection{Integration with DPTS Search Loop}

SSDP clustering is applied immediately after node expansion and PRM scoring:

\begin{algorithmic}[1]
\While{$t < T_{\max}$ and stopping criteria not met}
    \State $\mathcal{S} \leftarrow$ \textsc{Select}($\mathcal{N}_{\text{all}}, k$) \Comment{UCB-based selection}
    \For{each node $n \in \mathcal{S}$}
        \State $\mathcal{C}_n \leftarrow$ \textsc{Expand}($n, b$) \Comment{Generate $b$ children}
        \For{each child $c \in \mathcal{C}_n$}
            \State $\phi(c) \leftarrow$ \textsc{PRM}($c$) \Comment{Score with reward model}
            \State $s(c) \leftarrow \mathcal{E}(\text{decode}(x_c))$ \Comment{Compute embedding}
        \EndFor
        \State $\mathcal{C}'_n \leftarrow$ \textsc{ClusterAndPrune}($\mathcal{C}_n, \tau$) \Comment{SSDP merging}
        \State $n.\text{children} \leftarrow \mathcal{C}'_n$
    \EndFor
    \State $\mathcal{N}_{\text{all}} \leftarrow \mathcal{N}_{\text{all}} \cup \bigcup_{n \in \mathcal{S}} \mathcal{C}'_n$
    \State \textsc{Backpropagate}($\mathcal{S}$)
\EndWhile
\end{algorithmic}

\subsection{Early Stopping and Pruning Mechanisms}

SSDP inherits DPTS's adaptive stopping criteria to prevent over-exploration:

\textbf{Early Stopping:} During rollout, if a node's reward $\phi(n)$ falls below a threshold 
$\theta_{\text{es}} = \lambda_{\text{es}} \cdot \bar{\phi}_{\text{explore}}$, where 
$\bar{\phi}_{\text{explore}}$ is the mean reward of explored nodes and $\lambda_{\text{es}} = 0.8$, 
the rollout terminates early.

\textbf{Deep Seek Threshold:} For exploration nodes, a stricter threshold 
$\theta_{\text{ds}} = \lambda_{\text{ds}} \cdot \bar{\phi}_{\text{explore}}$ (with $\lambda_{\text{ds}} = 0.8$) 
is applied to prune low-quality branches.

\textbf{Solution Quality Gating:} Once $t^* = 5$ solutions are found, new rollouts are only pursued if 
the selected node's reward exceeds the best known solution reward: 
$\phi(n) > \max_{n \in \mathcal{N}_{\text{term}}} \phi(n)$.

\textbf{Time-Based Stopping:} The search terminates when wall-clock time exceeds $T_{\max}$ 
(default: 120 seconds) or after a maximum number of rollouts (default: 20).

\newpage

\subsection{Hyperparameters}

\FloatBarrier
\begin{table}[h]
\centering
\caption{SSDP Hyperparameters}
\begin{tabular}{lcc}
\hline
\textbf{Parameter} & \textbf{Symbol} & \textbf{Default Value} \\
\hline
Similarity threshold & $\tau$ & 0.75 \\
Tree width (beam size) & $b$ & 4 \\
Exploration weight & $w$ & $1/\sqrt{2}$ \\
Early stopping factor & $\lambda_{\text{es}}$ & 0.8 \\
Deep seek factor & $\lambda_{\text{ds}}$ & 0.8 \\
Solution count threshold & $t^*$ & 5 \\
Max search time & $T_{\max}$ & 120s \\
Max rollouts & $R_{\max}$ & 20 \\
Exploit ratio & $p$ & 0.5 \\
\hline
\end{tabular}
\end{table}
\FloatBarrier

\subsection{Embedding Models}
We evaluate the following sentence transformer architecture:
\begin{itemize}
    \setlength{\itemsep}{0pt}
    \setlength{\parskip}{0pt}
    \item \texttt{all-MiniLM-L6-v2}: 22M parameters, 384-dim embeddings, fast inference

\end{itemize}
The model is frozen during inference and kept on GPU for low-latency encoding.

\subsection{Merging Algorithm}

\begin{nolinenumbers}   

\begin{algorithm}[H]
\caption{\textsc{MergeCluster}($C$)}
\label{alg:mergecluster}
\begin{algorithmic}[1]
\Require cluster $C = \{c_1,\dots,c_m\}$
\State \textbf{Option A:} choose representative $r \leftarrow \arg\max_{c\in C} S(c)$
\State \textbf{Option B:} compute centroid embedding
\[
    \bar{e} \leftarrow \frac{1}{|C|}\sum_{c\in C} e(c)
\]
and create merged node from $\bar{e}$
\State Recompute $S(r)$ (e.g., rescore or use representative's score)
\State \Return $r$
\end{algorithmic}
\end{algorithm}

\end{nolinenumbers}  

\FloatBarrier    


\subsection{Reward model and prompts}
All methods use the same system prompt template (instructing stepwise chain-of-thought output and final-answer extraction). Candidate traces are rescored using the Math-Shepherd reward model \texttt{peiyi9979/math-shepherd-mistral-7b-prm} for reranking and final answer selection. Rescoring with an external reward model ensures consistent selection criteria across SSDP, DPTS, Best-of-N, Beam Search, and MCTS.

\newpage



\subsection{SSDP Method Example}
\FloatBarrier

\begin{figure}[htbp]
\centering
\begin{tikzpicture}[
  node/.style={rectangle,draw=black,rounded corners=4pt,minimum width=9mm,minimum height=9mm,inner sep=1pt,align=center,font=\small},
  mergednode/.style={rectangle,draw=black,rounded corners=4pt,minimum width=9mm,minimum height=9mm,inner sep=1pt,align=center,fill=yellow!35,font=\small},
  greennode/.style={rectangle,draw=black,rounded corners=4pt,minimum width=9mm,minimum height=9mm,inner sep=1pt,align=center,fill=green!40,font=\small},
  >=Stealth,
  arrow/.style={->,semithick},
  score/.style={font=\scriptsize, fill=white, inner sep=0.6pt} 
]

\node[node] (n0) at (0,0) {0};

\node[node] (n1) at (2,1) {1};
\node[node] (n2) at (2,-1) {2};

\node[node] (n5) at (4,2) {3};
\node[node] (n6) at (4,1) {4};
\node[node] (n3) at (4,0) {5};
\node[node] (n4) at (6,-2) {6};

\node[mergednode] (n11) at (6,1) {M5};

\node[node] (n7)  at (8, -1) {7};
\node[node] (n8)  at (8, -3) {8};
\node[mergednode] (n9)  at (10, -2) {M8};

\node[node] (n10) at (10,1) {9};

\node[node] (n12) at (12,1) {10};
\node[greennode] (n13) at (12,-1) {11}; 
\node[node] (n14) at (12,-3) {12};

\draw[arrow] (n0) -- node[midway, above, score] {0.90} (n1);
\draw[arrow] (n0) -- node[midway, below, score] {0.85} (n2);

\draw[arrow] (n1) -- node[midway, above, score] {0.88} (n5);
\draw[arrow] (n1) -- node[midway, above, score] {0.82} (n6);
\draw[arrow] (n1) -- node[midway, below, score] {0.80} (n3);

\draw[arrow] (n2) -- node[midway, below, score] {0.78} (n4);

\draw[arrow] (n5) -- node[midway, above, score] {0.76} (n11);
\draw[arrow] (n6) -- node[midway, above, score] {0.79} (n11);
\draw[arrow] (n3) -- node[midway, below, score] {0.74} (n11);

\draw[arrow] (n4) -- node[midway, above, score] {0.70} (n7);
\draw[arrow] (n4) -- node[midway, below, score] {0.68} (n8);

\draw[arrow] (n11) -- node[midway, above, score] {0.83} (n10);

\draw[arrow] (n7) -- node[midway, above, score] {0.71} (n9);
\draw[arrow] (n8) -- node[midway, below, score] {0.69} (n9);

\draw[arrow] (n10) -- node[midway, above, score] {0.60} (n12);
\draw[arrow] (n9) -- node[midway, below, score] {0.87} (n13);

\draw[arrow] (n9) -- node[midway, below, score] {0.65} (n14);

\node[draw=black, very thick, rounded corners=6pt, inner sep=4pt, fit=(n5)(n6)(n3)(n11), label=above:{\small Merged cluster A}] (mb1) {};
\node[draw=black, very thick, rounded corners=6pt, inner sep=4pt, fit=(n7)(n8)(n9),  label=above:{\small Merged cluster B}] (mb2) {};

\end{tikzpicture}
\end{figure}

\FloatBarrier

\FloatBarrier
\begingroup
\setlength{\tabcolsep}{10pt}    
\renewcommand{\arraystretch}{1.45} 

\begin{table}[htbp]
\centering
\small
\caption{Node numbers and short description of each reasoning step shown in the SSDP figure. Hypothetical reward model scores are attached to the arrows between nodes to simulate an end-to-end SSDP process.}
\begin{tabular}{|l|>{\raggedright\arraybackslash}p{0.9\textwidth}|}
\hline
\textbf{Node \#} & \textbf{Reasoning step} \\ \hline
0 & Problem statement: Find positive integer pairs \((x,y)\) satisfying \(\dfrac{1}{x}+\dfrac{1}{y}=\dfrac{1}{6}\) and whose sum is the smallest possible odd number. \\ \hline
1 & {Branch S1 — Algebraic rewrite:} Rearrange to expose factorization structure, e.g. \((x-6)(y-6)=36\). \\ \hline
2 & {Branch S2 — Divisor/substitution path:} Express \(y\) in terms of \(x\) via \(y=\dfrac{6x}{x-6}\) and enumerate divisors. \\ \hline
3 & {S1 child (factor pair 1):} Factor \(36 = 1\times 36\) \(\Rightarrow\) candidate \((7,42)\) and \((9,18)\) and \((8,24)\). \\ \hline
4 & {S1 child (factor pair 2):} Factor \(36 = 2\times 18\) \(\Rightarrow\) candidate \((8,24)\) and \((7,42)\) and \((9,18)\). \\ \hline
5 & {S1 child (factor pair 3):} Factor \(36 = 3\times 12\) \(\Rightarrow\) candidate \((9,18)\) and \((8,24)\) and \((7,42)\). \\ \hline
M5 & {S1 child (factor pair 3):} Factor \(36 = 3\times 12\) \(\Rightarrow\) candidate \((9,18)\) and \((8,24)\) and \((7,42)\). \\ \hline
6 & Performing isolation for y to then extract (x,y) values. \\ \hline
7 & {S2 child (divisor trial 1):} Divisor enumeration yields candidate \((12,12)\) and \((15,15)\). \\ \hline
8 & {S2 child (divisor trial 2):} Divisor enumeration yields candidate \((10,15)\) and \((12,12)\). \\ \hline
M8 & {S2 child (divisor trial 2):} Divisor enumeration yields candidate \((10,15)\) and \((12,12)\). \\ \hline
9 & {Find odd sums:} Add them both together, then modulo 2 to see if =1.\\ \hline
10 & \textbf{Incorrect Answer}: (9,18) and (7,42). This is incorrect because we are asking for the smallest sum, and both are incorrect - there is a smaller combination. \\ \hline
11 & \textbf{Answer found:} (10,15) because only one of them is odd, and no numbers between 10 and 15 were found that can be used. \\ \hline
12 & {Reasoning still in process}: Calculating the sums of 10+15 and 12+12. \\ \hline
\end{tabular}


\label{tab:node-steps-grid}
\end{table}

\endgroup
\FloatBarrier

\newpage

\end{document}